\documentclass[10pt,english,oneside,twocolumn,a4paper]{article}

\usepackage{graphicx}
\usepackage[utf8]{inputenc}
\usepackage{mathptmx}
\usepackage{tabularx}
\usepackage{ragged2e}
\usepackage[singlelinecheck=off]{caption}
\usepackage[T1]{fontenc}
\usepackage[numbers]{natbib}
\usepackage{amsmath}
\usepackage{tikz}
\usepackage{pgfplots}
\usepackage{multirow}

\usepackage{babel}
\RequirePackage[blocks]{authblk}
\makeatletter
\let\ps@plain\ps@empty
\def\@xivpt{14pt}
\def\@sect#1#2#3#4#5#6[#7]#8{%
  \ifnum #2<2
    \null\par\vskip-15pt
  \fi
  \ifnum #2>\c@secnumdepth 
    \let\@svsec\@empty
  \else
    \refstepcounter{#1}%
    \protected@edef\@svsec{%
      \ifnum #2<4
        \hb@xt@10mm{\csname the#1\endcsname}\relax
      \else
        \hb@xt@12mm{\csname the#1\endcsname}\relax
      \fi}%
  \fi
  \@tempskipa #5\relax
  \ifdim \@tempskipa>\z@
    \begingroup
      #6{%
        \@hangfrom{\hskip #3\relax\@svsec}%
          \interlinepenalty \@M #8\@@par}%
    \endgroup
    \csname #1mark\endcsname{#7}%
    \addcontentsline{toc}{#1}{%
      \ifnum #2>\c@secnumdepth \else  
        \protect\numberline{\csname the#1\endcsname}%
      \fi 
      #7}%
  \else
    \def\@svsechd{%
      #6{\hskip #3\relax
      \@svsec #8}%
      \csname #1mark\endcsname{#7}%
      \addcontentsline{toc}{#1}{%
        \ifnum #2>\c@secnumdepth \else
          \protect\numberline{\csname the#1\endcsname}%
        \fi
        #7}}%
  \fi
  \@xsect{#5}}
\renewcommand\LARGE{\@setfontsize\LARGE{16}{20}}
\def\abstract#1{\def\@abstract{#1}}
\def\abstractEn#1{\def\@abstractEn{#1}}
\def\titleEn#1{\def\@titleEn{#1}}
\def\@maketitle{%
  \newpage
  \null
  \let \footnote \thanks
    {\LARGE\bfseries\RaggedRight \@title \par}%
    \vskip 1\baselineskip%
    {\normalsize
      \@author\par}%
    \vskip 2\baselineskip%
    \vskip \baselineskip%
    {\section*{Abstract}
      \@abstract}%
  \par
  \vskip 3\baselineskip}

\renewcommand\section{\@startsection {section}{1}{\z@}%
                                   {-3.5ex \@plus -1ex \@minus -.2ex}%
                                   {\baselineskip}%
                                   {\normalfont\Large\bfseries\RaggedRight}}
\renewcommand\subsection{\@startsection{subsection}{2}{\z@}%
                                     {\baselineskip}%
                                     {1ex}%
                                     {\normalfont\large\bfseries\RaggedRight}}
\renewcommand\subsubsection{\@startsection{subsubsection}{3}{\z@}%
                                     {1\baselineskip}%
                                     {3bp}%
                                     {\normalfont\normalsize\bfseries\RaggedRight}}
\renewcommand\paragraph{\@startsection{paragraph}{4}{\z@}%
                                    {1\baselineskip\@plus1ex \@minus.2ex}%
                                    {3bp}%
                                    {\normalfont\normalsize\RaggedRight}}
\renewcommand\subparagraph{\@startsection{subparagraph}{5}{\parindent}%
                                       {3.25ex \@plus1ex \@minus .2ex}%
                                       {-1em}%
                                      {\normalfont\normalsize\bfseries\RaggedRight}}
\parindent\p@
\parskip\z@
\DeclareCaptionLabelSeparator{enskip}{\enskip}
\makeatother

\usepackage[hyphens]{url}
\usepackage{subfig}

\usepackage{todonotes}
\usepackage{siunitx}
\usepackage{array}
\usepackage{csquotes}
\usepackage{booktabs}
\usepackage{microtype}

\usepackage{todonotes}

\usepgfplotslibrary[groupplots]

\title{\mbox{Ego-Motion Estimation and Dynamic Motion Separation from 3D Point} Clouds for Accumulating Data and Improving 3D Object Detection}
\author{Patrick Palmer, TU Dortmund, Institute of Control Theory and Systems Engineering, 44227 Dortmund,\\ patrick.palmer@tu-dortmund.de\\
	Martin Krüger, TU Dortmund, Institute of Control Theory and Systems Engineering, 44227 Dortmund,\\ martin2.krueger@tu-dortmund.de\\
	Dr. Richard Altendorfer, ZF Group, 56070 Koblenz, richard.altendorfer@zf.com\\
	Univ.-Prof. Dr.-Ing. Prof. h.c. Dr. h.c. Torsten Bertram, TU Dortmund, Institute of Control Theory and Systems Engineering, 44227 Dortmund, torsten.bertram@tu-dortmund.de \vspace{-0.25cm}}

\abstract{
New 3+1D high-resolution radar sensors are gaining importance for 3D object detection in the automotive domain due to their relative affordability and improved detection compared to classic low-resolution radar sensors. One limitation of high-resolution radar sensors, compared to lidar sensors, is the sparsity of the generated point cloud. This sparsity could be partially overcome by accumulating radar point clouds of subsequent time steps. This contribution analyzes limitations of accumulating radar point clouds on the View-of-Delft dataset \cite{ViewOfDelftDataset_22}. By employing different ego-motion estimation approaches, the dataset's inherent constraints, and possible solutions are analyzed. Additionally, a learning-based instance motion estimation approach is deployed to investigate the influence of dynamic motion on the accumulated point cloud for object detection. Experiments document an improved object detection performance by applying an ego-motion estimation and dynamic motion correction approach.}

\begin{document}

\maketitle

\section{Introduction}
\label{sec:introduction}
\begin{figure*}[!htbp]
	\centering
	\includegraphics[height=4.9cm]{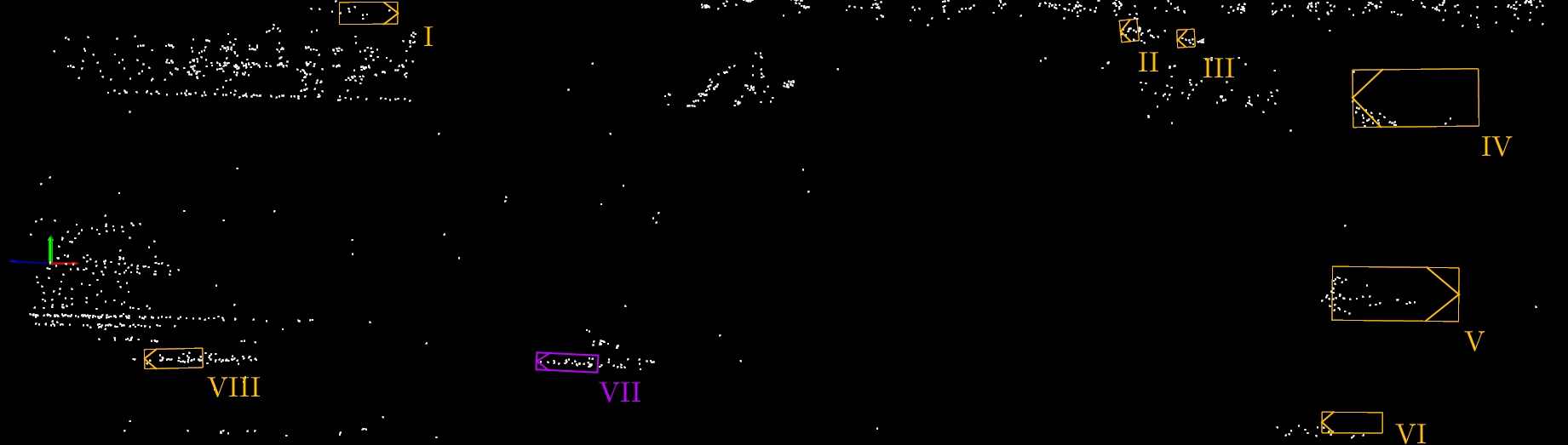}
	\caption{Bird's eye view perspective of a traffic scene visualizing an aggregated point cloud, accumulated over five frames. The ego-vehicle is located on the left, represented by the colored coordinate system indicating the sensor mounting position (viewing towards the right). Orange boxes represent annotated ground truth bounding boxes and the purple rectangle indicates a manually added ground truth bounding box. The roman numerals mark the objects identifier (ID). The arrow represents the direction of motion.}
	\label{Example_label}
	\vspace{-0.25cm}
\end{figure*}

One of the critical challenges of automating vehicles and the driving process is the perception of the environment. Precise knowledge of the traffic scene is necessary to make well-informed decisions on the automated vehicle's path planning and react adequately to sudden actions of other traffic participants. False, missing or imprecise detections entail errors in the environment model. This increases the risk of accidents and limits the time horizon for safe and comfortable vehicle path planning. Different sensor modalities are utilized in research and production vehicles for environment perception. Research vehicles mainly use high-resolution lidars due to their high information density and accuracy, whereas series production vehicles primarily use low-cost sensors like radars and cameras. An emerging sensor technology that is supposed to bridge the gap between lidar and traditional low-resolution radar sensors are 3+1D high-resolution radar sensors. Compared to traditional radars, these high-resolution radar sensors also measure the elevation angle and generate a denser point cloud while preserving the advantages of radar sensors like the direct estimation of the relative radial velocity $v_{rr}$, robustness against adverse weather conditions, and relatively low cost. Despite improvements in radar technology, currently available 3+1D radar sensors still suffer from noisy measurements and relatively sparse point clouds (compared to lidar). These limitations constrain the perception performance and thus affect the following modules (and their performance) in an automated driving stack. A common strategy to overcome the sparsity of low-resolution lidar and radar point clouds is the aggregation of information over concurrent time steps by accumulation. This yields a denser point cloud that can be used for perception tasks like 3D object detection. 
For static objects, an accumulation can be done by transforming point positions from the previous to the current coordinate frame using the ego-motion alone. Nowadays, ego-motion is mostly estimated by combining GPS, wheel odometry, and inertial measurements (angular rates and accelerations) and is available for most public datasets. Targets from dynamic objects, such as reflections from cars, pedestrians, and cyclists, must be considered separately. A naive accumulation of these points based on the ego-motion alone results in an error, represented by trailing points behind the object. An example can be seen in Fig. \ref{Example_label}, the bicycle on the bottom left (ID: VIII) has a tail of points outside the bounding box. Due to the radar sensor's direct measurement of radar radial velocity $v_{rr}$, an accurate distinction between static and dynamic points is possible \cite{RadarSegmentation_18}. The $v_{rr}$ can additionally be used for estimating the velocity over ground of objects. Considering the motion of dynamic objects when accumulating point clouds from subsequent frames leads to improved consistency of the accumulated point cloud. Hence, the accuracy of object detection approaches applied to the point cloud increases \cite{FullVelocityRadarLong2021}.

\vspace{0.1cm}

\hspace{0.3cm} \textbf{Related Work:}
For lidar \cite{RigitSeneFlow_Dewan_2016, Flownet3d_Liu_2019} and 3+1D high-resolution radar data \cite{SSSS4DR_Liu_2022} flow-based methods, that only utilize the point coordinates, can accurately estimate the scene flow for a point cloud. The point cloud can be separated into static and dynamic areas using the scene flow. Static points can then be used to estimate the ego-motion, while the instance's motion can be derived from the dynamic points. By extending the considered time horizon and employing motion segmentation \cite{PointCloudAccumulation_22} (PCAc) has shown that accurate accumulation of point clouds is possible given only the naively accumulated point clouds.

As summarized by \cite{TowarddeepradarZhou2022}, multiple 3+1D radar datasets have been proposed, each suffering from unique limitations, making most of them unfeasible for the ultimate goal of 3D object detection. The View-of-Delft (VoD) dataset \cite{ViewOfDelftDataset_22} meets the requirements. It provides time-continuous labeled objects, non-accumulated radar data, and additional sensor modalities like high-resolution lidar and camera sensor data.

The main contributions of the work at hand are:
\begin{itemize}
	\item Evaluating ego-motion estimation approaches to accumulate high-resolution radar point clouds on subsequent frames on the VoD dataset.
	\vspace{-0.15cm}
	\item Adaptation of PCAc \cite{PointCloudAccumulation_22} approach to the VoD dataset and evaluation of the transferability to radar data.
	\vspace{-0.15cm}
	\item Investigating the influence of different dynamic and static motion correction approaches on the performance of 3D object detection.
\end{itemize}
\vspace{-0.25cm}

\section{Limitations of the VoD dataset}
\vspace{-0.15cm}
The VoD dataset is one of very few datasets that include 3+1D high-resolution radar measurements. In addition to the radar sensor, mono camera and high-resolution 64-layer lidar data are provided. Due to the dataset's focus on urban scenarios where different types of traffic participants share the same space, only a limited view range of up to 51.2m is considered, and the amount of observed pedestrians and cyclist is high compared to other datasets like the Astyx dataset \cite{AstyxMeyer2019}.
One major limitation of the dataset is the accuracy of labeled objects. Multiple objects are either not labeled or mislabeled. An example is visualized in Fig. \ref{Example_label}. In the lower part of the image are three bicycles. The foremost one (ID: VIII) is labeled correctly. The middle one (ID: VII) is not labeled at all, and the label for the sternmost one (ID: VI) is not well aligned to the cluster of close targets. The questionable quality of such annotations negatively influences the training because misaligned or missing annotations confuse the training procedure. Due to the time and effort necessary for correcting annotations, only minor label adaptations have been applied to mitigate the problem of object reflection points that lay outside the bounding box. The box is extended by \SI{0.2}{\meter} in each direction to capture most of the points belonging to an object, similar to \cite{VoD_Baratam_2022}.

The second observed limitation is the unavailability of ground truth ego-motion information. The authors provide an accumulated set of radar scans but do not provide the utilized ego-motion for generating them. This limits the ability to investigate dynamic object motion. Additionally, the reproduction of the accumulated scans is hindered by different sampling frequencies of the radar sensor and the provided single radar scans. The utilized sensor captures the scene with a frequency of \SI{13}{\hertz}, but the single-frame data is only provided with a frequency of \SI{10}{\hertz}, to synchronize the measurement of the radar to the other sensor modalities. The provided accumulated point cloud is collected from the base \SI{13}{\hertz} data. Certain radar scans are therefore not provided and the time interval between the provided frames varies. Different strategies of estimating ego-motion are investigated in the following chapter to mitigate the missing ego-motion information.

\subsection*{Ego-Motion Estimation}
\label{sec:em_estimation}
The ego-vehicle motion is described as a transformation matrix that consists of the translation and rotation of the ego-vehicle between two consecutive time steps. It can be derived either from an inertial measurement-based system or directly from environmental measurement sensors.

\hspace{0.3cm} \textbf{Ego-Motion-Estimation from Pose (EM-P):} The VoD dataset does not directly provide information about the ego-motion. However, transformation matrices from the camera coordinate system to a \textit{map} and a \textit{world} coordinate system are provided. Both transformations only differ in the definition of the point of origin. Hence, the ego-motion is calculated from the transformation between consecutive frames.

\hspace{0.3cm} \textbf{Ego-Motion-Estimation from Point Cloud with GICP (EM-G):} Due to limitations observed in the ego-motion from the VoD pose estimation, a method for estimating the motion directly from the recorded point cloud is investigated. Generalized Iterative Closest Point  (GICP)\cite{GenerelizedICPSegal2009} is chosen due to its simple implementation and the assumed sufficient accuracy. It is applied to the lidar and high-resolution radar point cloud separately.

\begin{figure*}[htbp!]
	\begin{tikzpicture}
	
	\definecolor{crimson2143940}{RGB}{214,39,40}
	\definecolor{darkgray176}{RGB}{176,176,176}
	\definecolor{darkorange25512714}{RGB}{250,127,14}
	\definecolor{goldenrod18818934}{RGB}{188,189,34}
	\definecolor{lightgray204}{RGB}{204,204,204}
	\definecolor{orchid227119194}{RGB}{227,119,194}
	\definecolor{sienna1408675}{RGB}{140,86,75}
	\definecolor{steelblue31119180}{RGB}{31,119,180}
	
	\begin{groupplot}[group style={group size=1 by 2}]
	\nextgroupplot[
	scaled x ticks=manual:{}{\pgfmathparse{#1}},
	tick align=outside,
	tick pos=left,
	x grid style={darkgray176},
	xmin=0, xmax=80,
	xtick style={color=black},
	xtick={0,10,20,30,40,50,60,70,80},
	xticklabels={},
	y grid style={darkgray176},
	ymin=0, ymax=1.15,
	ytick style={color=black},
	width=17cm,
	height=5cm,
	legend columns=2, 
	]
	\addplot [semithick, steelblue31119180]
	table {%
		0 0.438737004995346
		1 0.482691675424576
		2 0.311861932277679
		3 0.370293080806732
		4 0.490102380514145
		5 0.326135993003845
		6 0.445709407329559
		7 0.507838487625122
		8 0.333940356969833
		9 0.333616644144058
		10 0.551302373409271
		11 0.32874208688736
		12 0.487235248088837
		13 0.52157187461853
		14 0.226855367422104
		15 0.514003455638885
		16 0.522544145584106
		17 0.341984003782272
		18 0.36026918888092
		19 0.67466926574707
		20 0.358760952949524
		21 0.408590316772461
		22 0.518781065940857
		23 0.360252767801285
		24 0.494204074144363
		25 0.52462112903595
		26 0.354497700929642
		27 0.409573405981064
		28 0.543561279773712
		29 0.356363624334335
		30 0.473811894655228
		31 0.537757575511932
		32 0.369567215442657
		33 0.400851458311081
		34 0.51854544878006
		35 0.34404045343399
		36 0.472613722085953
		37 0.531243622303009
		38 0.339523911476135
		39 0.40635159611702
		40 0.521618485450745
		41 0.343442678451538
		42 0.487129151821136
		43 0.492300987243652
		44 0.353374779224396
		45 0.410861402750015
		46 0.519237518310547
		47 0.347586274147034
		48 0.461942702531815
		49 0.519015908241272
		50 0.232724323868752
		51 0.510417997837067
		52 0.528623998165131
		53 0.365054726600647
		54 0.351292371749878
		55 0.659743309020996
		56 0.358224451541901
		57 0.397471815347672
		58 0.535697758197784
		59 0.343606680631638
		60 0.358544707298279
		61 0.659031331539154
		62 0.343574374914169
		63 0.369526922702789
		64 0.515974104404449
		65 0.320015430450439
		66 0.469534128904343
		67 0.483701527118683
		68 0.336411237716675
		69 0.375445514917374
		70 0.517933070659637
		71 0.284629553556442
		72 0.5125532746315
		73 0.495237678289413
		74 0.218749925494194
		75 0.501073658466339
		76 0.510859191417694
		77 0.328988194465637
		78 0.324202865362167
		79 0.627051174640656
		80 0.342200577259064
		81 0.37770488858223
		82 0.498590260744095
		83 0.337650865316391
		84 0.468303829431534
		85 0.521901905536652
		86 0.200958222150803
		87 0.587650775909424
		88 0.434681326150894
	};
	\addlegendentry{EM-P}
	\addplot [semithick, darkorange25512714]
	table {%
		0 0.308295667171478
		1 0.601454317569733
		2 0.179390639066696
		3 0.492844581604004
		4 0.448994278907776
		5 0.352069884538651
		6 0.318117022514343
		7 0.632694363594055
		8 0.250521212816238
		9 0.396091222763062
		10 0.511471211910248
		11 0.369570732116699
		12 0.316370010375977
		13 0.675750494003296
		14 0.177175149321556
		15 0.545466423034668
		16 0.487723797559738
		17 0.366617977619171
		18 0.304111123085022
		19 0.655054748058319
		20 0.223269924521446
		21 0.426510989665985
		22 0.548253655433655
		23 0.413503110408783
		24 0.352493852376938
		25 0.669370949268341
		26 0.225630551576614
		27 0.492863744497299
		28 0.557389199733734
		29 0.282104253768921
		30 0.392441511154175
		31 0.643015742301941
		32 0.196905598044395
		33 0.506463885307312
		34 0.488820612430573
		35 0.392381310462952
		36 0.346157819032669
		37 0.651563823223114
		38 0.177975505590439
		39 0.534839689731598
		40 0.510829269886017
		41 0.324670821428299
		42 0.341211944818497
		43 0.633673131465912
		44 0.211988613009453
		45 0.453237503767014
		46 0.469633519649506
		47 0.398408859968185
		48 0.354155540466309
		49 0.626089453697205
		50 0.170404389500618
		51 0.538971543312073
		52 0.46179386973381
		53 0.390746176242828
		54 0.287357121706009
		55 0.702268421649933
		56 0.183927580714226
		57 0.567823350429535
		58 0.469581067562103
		59 0.361795365810394
		60 0.317285269498825
		61 0.671656906604767
		62 0.236024305224419
		63 0.458767861127853
		64 0.498919755220413
		65 0.348378330469131
		66 0.359789371490479
		67 0.599498271942139
		68 0.255947232246399
		69 0.45573365688324
		70 0.520000278949738
		71 0.243351265788078
		72 0.42014268040657
		73 0.562101483345032
		74 0.214477002620697
		75 0.470548361539841
		76 0.526326179504395
		77 0.316739410161972
		78 0.359401643276215
		79 0.573848485946655
		80 0.273656249046326
		81 0.424029320478439
		82 0.529921233654022
		83 0.289105325937271
		84 0.410454958677292
		85 0.573527038097382
		86 0.182072162628174
		87 0.532638549804688
		88 0.519073069095612
	};
	\addlegendentry{EM-G-Lidar}
	\addplot [semithick, orchid227119194]
	table {%
		0 0.385072827339172
		1 0.388866305351257
		2 0.41044819355011
		3 0.388051599264145
		4 0.393876284360886
		5 0.416394650936127
		6 0.397174894809723
		7 0.398811787366867
		8 0.404018461704254
		9 0.415873557329178
		10 0.399747997522354
		11 0.410160809755325
		12 0.413077622652054
		13 0.412786453962326
		14 0.419962465763092
		15 0.407738119363785
		16 0.432633996009827
		17 0.428676098585129
		18 0.428183972835541
		19 0.42614084482193
		20 0.4226915538311
		21 0.430373996496201
		22 0.410548090934753
		23 0.420636415481567
		24 0.428450584411621
		25 0.436514377593994
		26 0.438900411128998
		27 0.43929386138916
		28 0.450352638959885
		29 0.451875239610672
		30 0.429975420236588
		31 0.436633378267288
		32 0.432240843772888
		33 0.427453339099884
		34 0.429720044136047
		35 0.418291926383972
		36 0.436671435832977
		37 0.428957492113113
		38 0.430382192134857
		39 0.427227169275284
		40 0.431956470012665
		41 0.435624569654465
		42 0.42433950304985
		43 0.423515170812607
		44 0.420533388853073
		45 0.426202237606049
		46 0.412601888179779
		47 0.405735909938812
		48 0.418025583028793
		49 0.420182853937149
		50 0.418918907642365
		51 0.411988198757172
		52 0.426277220249176
		53 0.424970597028732
		54 0.423693478107452
		55 0.412560433149338
		56 0.425256907939911
		57 0.427510797977448
		58 0.432319402694702
		59 0.433617293834686
		60 0.428792148828506
		61 0.433780163526535
		62 0.428678244352341
		63 0.437361031770706
		64 0.419185131788254
		65 0.424074918031693
		66 0.421838730573654
		67 0.428922742605209
		68 0.416896313428879
		69 0.420216798782349
		70 0.419711112976074
		71 0.423224538564682
		72 0.40572002530098
		73 0.415778905153275
		74 0.409546107053757
		75 0.402634382247925
		76 0.405103504657745
		77 0.406157821416855
		78 0.418389171361923
		79 0.408265680074692
		80 0.410223513841629
		81 0.420086711645126
		82 0.412333548069
		83 0.412932723760605
		84 0.408327043056488
		85 0.416835933923721
		86 0.416782349348068
		87 0.40151834487915
		88 0.419619858264923
	};
	\addlegendentry{EM-mG-Lidar}
	\addplot [semithick, goldenrod18818934]
	table {%
		0 0.196375414729118
		1 0.762117385864258
		2 0.356458276510239
		3 0.393200218677521
		4 0.247606426477432
		5 0.257471978664398
		6 0.47940930724144
		7 0.667328238487244
		8 0.36792865395546
		9 0.601301252841949
		10 0.476645082235336
		11 0.097546398639679
		12 0.358199685811996
		13 0.702467560768127
		14 0.394757807254791
		15 0.305864304304123
		16 0.408003151416779
		17 0.144518151879311
		18 0.276576012372971
		19 0.55134505033493
		20 0.135304763913155
		21 0.643554866313934
		22 0.649587690830231
		23 0.481476068496704
		24 0.32883819937706
		25 0.501363337039948
		26 0.505843758583069
		27 0.487398266792297
		28 0.634946882724762
		29 0.209885463118553
		30 0.738441109657288
		31 0.807961940765381
		32 0.337609142065048
		33 0.565801978111267
		34 1.03200924396515
		35 0.171402454376221
		36 0.423925876617432
		37 0.874753773212433
		38 0.0162522383034229
		39 0.405114382505417
		40 0.684221386909485
		41 0.510778903961182
		42 0.469457447528839
		43 0.511534988880157
		44 0.321034610271454
		45 0.397480547428131
		46 0.215356722474098
		47 0.397854030132294
		48 0.455625623464584
		49 0.495200902223587
		50 0.085988387465477
		51 0.0843495950102806
		52 0.230624675750732
		53 0.227472275495529
		54 0.57129967212677
		55 0.688384115695953
		56 0.426048368215561
		57 0.471077889204025
		58 0.45031750202179
		59 0.290564179420471
		60 0.218524843454361
		61 0.302643179893494
		62 0.488220721483231
		63 0.323363155126572
		64 0.519621610641479
		65 0.415458828210831
		66 0.453012943267822
		67 0.717390239238739
		68 0.319369196891785
		69 0.559856116771698
		70 0.443464696407318
		71 0.1771469861269
		72 0.255295187234879
		73 0.730904161930084
		74 0.0320485681295395
		75 0.388733208179474
		76 0.520624756813049
		77 0.46746352314949
		78 0.0920755714178085
		79 0.556308507919312
		80 0.436354428529739
		81 0.368369787931442
		82 0.0583481118083
		83 0.514176726341248
		84 0.324225634336472
		85 0.322175025939941
		86 0.355464547872543
		87 0.54231208562851
		88 0.48566797375679
	};
	\addlegendentry{EM-G-Radar}
	
	\nextgroupplot[
	tick align=outside,
	tick pos=left,
	x grid style={darkgray176},
	xlabel={Time [s]},
	xmin=0, xmax=80,
	xtick style={color=black},
	xtick={0,10,20,30,40,50,60,70,80},
	xticklabels={0,1,2,3,4,5,6,7,8},
	y grid style={darkgray176},
	ymin=0, ymax=1.15,
	ytick style={color=black},
	width=17cm,
	height=5cm,
	legend columns=2,
	]
	\addplot [semithick, steelblue31119180]
	table {%
		0 0.438737004995346
		1 0.482691675424576
		2 0.311861932277679
		3 0.370293080806732
		4 0.490102380514145
		5 0.326135993003845
		6 0.445709407329559
		7 0.507838487625122
		8 0.333940356969833
		9 0.333616644144058
		10 0.551302373409271
		11 0.32874208688736
		12 0.487235248088837
		13 0.52157187461853
		14 0.226855367422104
		15 0.514003455638885
		16 0.522544145584106
		17 0.341984003782272
		18 0.36026918888092
		19 0.67466926574707
		20 0.358760952949524
		21 0.408590316772461
		22 0.518781065940857
		23 0.360252767801285
		24 0.494204074144363
		25 0.52462112903595
		26 0.354497700929642
		27 0.409573405981064
		28 0.543561279773712
		29 0.356363624334335
		30 0.473811894655228
		31 0.537757575511932
		32 0.369567215442657
		33 0.400851458311081
		34 0.51854544878006
		35 0.34404045343399
		36 0.472613722085953
		37 0.531243622303009
		38 0.339523911476135
		39 0.40635159611702
		40 0.521618485450745
		41 0.343442678451538
		42 0.487129151821136
		43 0.492300987243652
		44 0.353374779224396
		45 0.410861402750015
		46 0.519237518310547
		47 0.347586274147034
		48 0.461942702531815
		49 0.519015908241272
		50 0.232724323868752
		51 0.510417997837067
		52 0.528623998165131
		53 0.365054726600647
		54 0.351292371749878
		55 0.659743309020996
		56 0.358224451541901
		57 0.397471815347672
		58 0.535697758197784
		59 0.343606680631638
		60 0.358544707298279
		61 0.659031331539154
		62 0.343574374914169
		63 0.369526922702789
		64 0.515974104404449
		65 0.320015430450439
		66 0.469534128904343
		67 0.483701527118683
		68 0.336411237716675
		69 0.375445514917374
		70 0.517933070659637
		71 0.284629553556442
		72 0.5125532746315
		73 0.495237678289413
		74 0.218749925494194
		75 0.501073658466339
		76 0.510859191417694
		77 0.328988194465637
		78 0.324202865362167
		79 0.627051174640656
		80 0.342200577259064
		81 0.37770488858223
		82 0.498590260744095
		83 0.337650865316391
		84 0.468303829431534
		85 0.521901905536652
		86 0.200958222150803
		87 0.587650775909424
		88 0.434681326150894
	};
	\addlegendentry{EM-P}
	\addplot [semithick, sienna1408675]
	table {%
		0 0.355294945200838
		0 0.388380845446043
		1 0.356975062856848
		2 0.444098933304985
		3 0.345975580804514
		4 0.407358066860403
		5 0.339773752742801
		6 0.386457413561979
		7 0.384244874864042
		8 0.446882574191053
		9 0.316112336914755
		10 0.449498878619489
		11 0.343060399965143
		12 0.47236640863667
		13 0.331680322705524
		14 0.459866408989089
		15 0.344762764910287
		16 0.563664418669078
		17 0.224847694105433
		18 0.467223843810555
		19 0.494191385380161
		20 0.327895866797626
		21 0.360650386250308
		22 0.610428677683681
		23 0.2593472045897
		24 0.405815028988702
		25 0.569245216972071
		26 0.298964002219896
		27 0.378007686718581
		28 0.655891539317721
		29 0.216890978458705
		30 0.511481802604011
		31 0.570528803063737
		32 0.287098218904647
		33 0.371400449496379
		34 0.677629931510739
		35 0.228385145599506
		36 0.485168012760103
		37 0.582054345738137
		38 0.321641359787024
		39 0.405915298625198
		40 0.637872637870106
		41 0.241220148846403
		42 0.437130036925437
		43 0.642978883920974
		44 0.247501950844325
		45 0.418565486316402
		46 0.671269464165161
		47 0.262377726162806
		48 0.447904398274784
		49 0.588296451105441
		50 0.280458240454985
		51 0.414486372195459
		52 0.60814097869277
		53 0.279275843764678
		54 0.519499890978816
		55 0.561257201346863
		56 0.358841740383462
		57 0.337861048976834
		58 0.218203038659034
		59 0.195825102276838
		60 0.48057590025554
		61 0.591684437469165
		62 0.290828600555141
		63 0.410366530192466
		64 0.618759838653862
		65 0.21576083932526
		66 0.480863142551296
		67 0.591131942286976
		68 0.291907649647824
		69 0.368465931557662
		70 0.672120624756504
		71 0.257174416868267
		72 0.484034026987859
		73 0.57922397425648
		74 0.26216589023917
		75 0.39525992520133
		76 0.668458224068435
		77 0.214614254461578
		78 0.464169231745263
		79 0.581923446906321
		80 0.26702736737896
		81 0.39154321926936
		82 0.311336709611594
		83 0.246608447313437
		84 0.413289146981574
		85 0.521305430554364
		86 0.301714644984032
		87 0.464282040763221
		88 0.483739277074505
		89 0.166207904301998
		90 0.213792598222593
		91 0.49590161178534
		92 0.345862850219364
		93 0.137774266471475
		94 0.580258862031226
		95 0.333038662866164
		96 0.373666107035441
		97 0.643330129849263
		98 0.278704966496922
		99 0.0494475174997627
		100 0.199756223050601
		101 0.117592076981738
		102 0.0106967798570818
		103 0.0720102271903762
		104 0.0118553597277585
		105 0.476924747496114
		106 0.00588867751816518
		107 0.00267717870565061
		108 0.00267717870565061
	};
	\addlegendentry{EM-SO}
	\addplot [semithick, crimson2143940]
	table {%
		
		0 0.403097778558731
		1 0.405827909708023
		2 0.407372295856476
		3 0.409867823123932
		4 0.412694752216339
		5 0.416618347167969
		6 0.417982488870621
		7 0.421485513448715
		8 0.423352897167206
		9 0.426214516162872
		10 0.427875846624374
		11 0.429806649684906
		12 0.432849824428558
		13 0.436700195074081
		14 0.438774347305298
		15 0.441542625427246
		16 0.44295272231102
		17 0.445674806833267
		18 0.44772869348526
		19 0.449918687343597
		20 0.450499385595322
		21 0.452474504709244
		22 0.451503187417984
		23 0.452711969614029
		24 0.452971756458282
		25 0.452974319458008
		26 0.453422784805298
		27 0.453881502151489
		28 0.453546047210693
		29 0.452313661575317
		30 0.45182079076767
		31 0.448661655187607
		32 0.448449075222015
		33 0.448517143726349
		34 0.445456981658936
		35 0.444286972284317
		36 0.444230377674103
		37 0.441477298736572
		38 0.441559970378876
		39 0.441065460443497
		40 0.439023345708847
		41 0.438290983438492
		42 0.438209056854248
		43 0.436072260141373
		44 0.435067474842072
		45 0.434954226016998
		46 0.43351349234581
		47 0.433944523334503
		48 0.435308873653412
		49 0.437043130397797
		50 0.437630653381348
		51 0.438002914190292
		52 0.439547836780548
		53 0.440385729074478
		54 0.442748010158539
		55 0.444399893283844
		56 0.44559234380722
		57 0.444894373416901
		58 0.441827774047852
		59 0.440891206264496
		60 0.439865171909332
		61 0.436483711004257
		62 0.433710873126984
		63 0.430930227041245
		64 0.427238881587982
		65 0.426547348499298
		66 0.426132261753082
		67 0.422510772943497
		68 0.421260505914688
		69 0.420232385396957
		70 0.419003009796143
		71 0.418726921081543
		72 0.420245081186295
		73 0.420700132846832
		74 0.421668589115143
		75 0.42172235250473
		76 0.420443475246429
		77 0.418703854084015
		78 0.418028503656387
		79 0.417199552059174
		80 0.418771833181381
		81 0.420352756977081
		82 0.421785205602646
		83 0.423302263021469
		84 0.424156486988068
		85 0.424421846866608
		86 0.424003124237061
		87 0.424469232559204
		88 0.424753099679947
	};
	\addlegendentry{EM-$v_{rr}$}
	\end{groupplot}

	\draw ({$(current bounding box.south west)!0.04!(current bounding box.south east)$}|-{$(current bounding box.south west)!0.5!(current bounding box.north west)$}) node[
	scale=1.1,
	anchor=west,
	text=black,
	rotate=90.0
	]at (-1,-5.2){Longitudinal translation between consecutive frames [m]};
	\end{tikzpicture}
	\caption{Translation between frames in driving direction estimated from various sensor modalities for an exemplary scene with an approximately constant velocity of \SI{15}{\kilo\meter\per\hour} in driving direction.}
	\label{tab:plot_em}
	\vspace{-0.25cm}
\end{figure*}
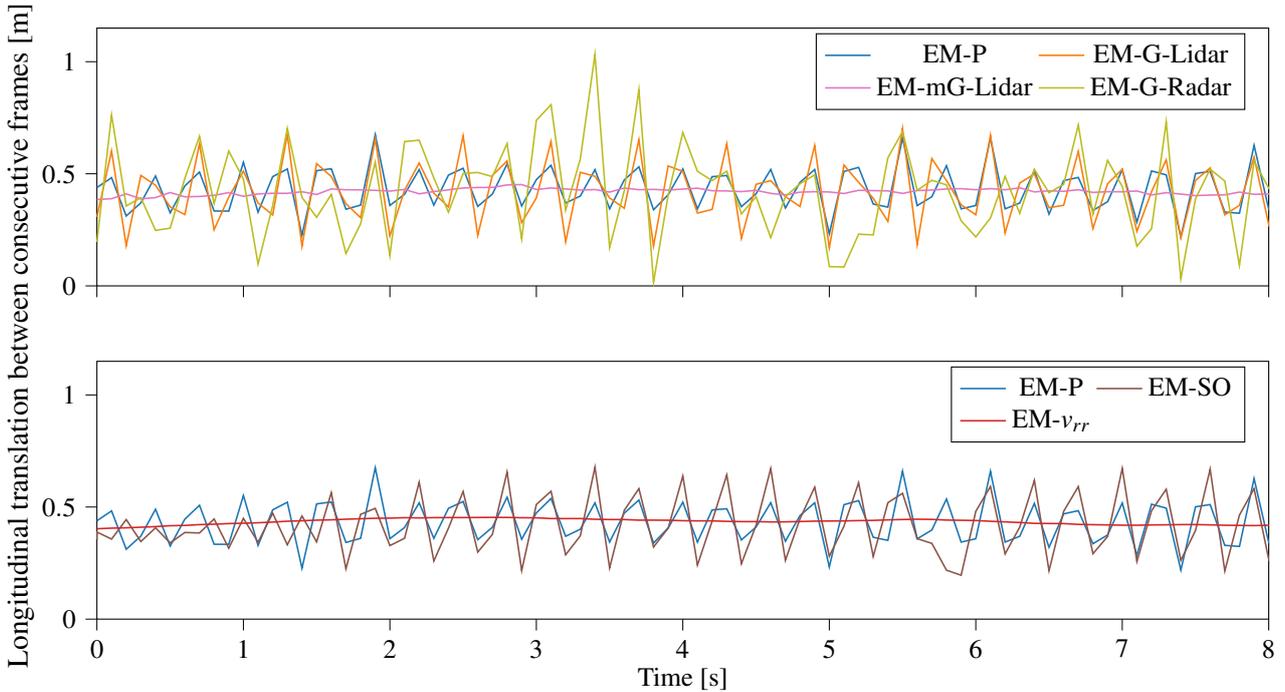

\hspace{0.3cm} \textbf{Averaging of estimated motion over multiple time steps (EM-mG):}
Due to the high oscillation of the EM-G output, as can be seen in Fig. \ref{tab:plot_em}, an averaging of the transformation matrix is employed to reduce implausible fluctuations. The average of the translation and rotation matrices are calculated separately. Due to an observed pattern that repeats every 6 time steps, the translation is averaged by calculating the moving average over the translation at 6 consecutive time steps. An averaging of the rotation matrix is not trivial. One way of averaging utilized in this work is to transform the rotation matrix into a quaternion representation. Then, the average quaternion is calculated using singular value decomposition (SVD) \cite{AveragingQuaternionsMarkley2007}, over the same 6 consecutive time steps. Afterward, the quaternion is re-transformed into the rotation matrix. 

\hspace{0.3cm} \textbf{Estimation from measured radar radial velocity (EM-$v_{rr}$):}
One unique feature of the radar point cloud that can be utilized to estimate ego-motion is the direct measurement of $v_{rr}$. 
The ego-motion can be estimated by applying a least squares-based approach to the static points in the scene, as described in \cite{InstanteniousEgoMotionKellner2013}. The main difference to alternative approaches is, that only measurements from a single frame are needed. The differentiation of static and dynamic points is done by estimating the static points using a random sample consensus algorithm \cite{RANSAC_Fischler_1981}. Due to the additional information about the elevation angle $\varepsilon$ measured by the 3+1D radar sensor, the observation angle $\varphi$ is utilized, instead of the azimuth angle $\vartheta$. Calculated using the spherical Pythagorean theorem:
\begin{equation}
\varphi = \text{cos}^{-1}(\text{cos}(\vartheta) \text{cos}(\varepsilon))
\end{equation}

\hspace{0.3cm} \textbf{Estimation from static objects (EM-SO):}
The VoD dataset includes labels and track identifiers for some static objects like \textit{bicycle racks} and \textit{unused bicycles}. The track identifiers enable the re-identification of objects in consecutive frames.
Assuming noise-free data, the rotation $R$ and translation $t$ can be derived by solving:

\begin{equation}
RC_{k-1} + t = C_{k}
\end{equation}

Where $C$ represents the center points of all objects at time step $k-1$ and $k$. Due to noise in the data, a least squares-based approach is used to derive the motion. Unfortunately, static objects do not occur in every frame. That limits the usefulness and accuracy of this method significantly and also makes it dependent on the accuracy of the annotation of static objects. Hence, this approach is only considered as a baseline for comparing the previous methods against.

\vspace{-0.15cm}
\section{Treatment of Dynamic Objects}
\label{sec:treatment_of_dynamic_objects}
\vspace{-0.15cm}
Due to the motion of certain objects an accumulation of the point cloud by just considering ego-motion results in an error for moving objects. For the foremost bicycle on the bottom left in Fig. \ref{Example_label}, this is represented by the smearing pattern of points behind the object. This negatively influences the training as objects get represented differently depending on their motion.

\hspace{0.3cm} \textbf{Estimation from the provided labels (Dyn-GT):}
For a baseline approach the motion of objects can be derived from the labels by estimating the transformation between the center points of objects. This is the best-case baseline to which the other approaches will be compared.

\hspace{0.3cm} \textbf{Relative Radial Velocity (Dyn-$v_{rr}$):}
If an object only moves in a radial direction to the radar sensor, the full motion of the object can be measured accurately. Unfortunately, objects rarely just move in radial direction, which limits this approach.
Nevertheless, a correction by $v_{rr}$ will result in a good motion correction for all vehicles that move in the same or opposite direction to the ego-vehicle.

\hspace{0.3cm} \textbf{Learning-Based Estimation of Dynamic Motion (Dyn-PCAc):} 
To directly estimate the dynamic object motion from the point cloud, PCAc
\cite{PointCloudAccumulation_22} is adapted to the VoD dataset. PCAc \cite{PointCloudAccumulation_22} is a supervised end-to-end learning-based algorithm that first estimates the ego-motion (EM-PCAc) of the vehicle by separating the scene into static and dynamic areas and employing robust point matching on the static points. Afterward, motion segmentation and instance association steps are executed to derive the instance motion and correct the point's position in the accumulated point cloud. When using the estimated ego-motion and the labels of dynamic objects, a ground truth motion set can be derived. This is used during training.

\vspace{0.1cm}

\vspace{-0.25cm}

\section{Experimental Evaluation}
\label{sec:evaluation}
For the overall goal of improving the object detection performance the evaluation of the accuracy of the ego-motion estimation as well as the instance motion assignment are analyzed by investigating their influence on the object detection performance. PointPillars \cite{Pointpillars} is used as the 3D object detection method as implemented in the OpenPCDet framework \cite{openpcdet2020}. Minor adaptations to the model configuration were made, as described by \cite{ViewOfDelftDataset_22}.
PointPillars is chosen over other object detectors due to its popularity for radar-based perception methods, see e.g. \cite{ViewOfDelftDataset_22}. In previous investigations \cite{Palmer_2023} it has shown to provide at least comparable, if not better detection performance than other detectors while also being computationally efficient.

\hspace{0.3cm} \textbf{Training Details:}
All experiments are carried out with three different reproducible model initializations to reduce performance dependencies on the initialization. The model is trained for at most 125 epochs utilizing a one-cycle learning rate scheduler \cite{OneCycle_Smith_2019}. Due to observations that the bulk of training progress happens at low learning rates, the learning rate scheduler is adapted to spend more time at lower learning rates while preserving the one-cycle character. For the VoD test set no labels are provided as the authors intend to provide an evaluation server later, similar to the KITTI dataset \cite{KITTI_Geiger_2013}. Hence, we used the validation set for evaluation.

\hspace{0.3cm} \textbf{Evaluation Metrics:}
Similar to KITTI \cite{KITTI_Geiger_2013} and \cite{ViewOfDelftDataset_22}, we use Average Precision (AP) as the primary performance metric but require a higher minimum intersection over union (IoU) of [70\%, 50\%, 50\%] in [x, y, z] direction for \textit{cars} and [50\%, 50\%, 50\%] for \textit{pedestrians} and \textit{cyclists}. This stricter requirement emphasizes the improved accuracy of the detection position by dynamic motion handling as well as the performance in object height detection by a more significant difference between 3D and 2D Bird's eye view (BEV) evaluation. The mean Average Precision (mAP) is the mean of the three class-wise APs. The results are specified for two different range regions, short-range (SR): evaluating all objects at distance of \SI{0}{}-\SI{30}{\meter} from the ego-vehicle and long-range (LR): evaluating all objects at distances \SI{>30}{\meter}.
The symmetric Chamfer Distance (sCD) quantifies the distance between (static) points after ego-motion correction for two consecutive time steps and is used as a metric to represent the error of estimated ego-motion. Due to the noisy radar point cloud, this metric is calculated utilizing the lidar point cloud.

\vspace{0.08cm}
\subsection{Ego-Motion Estimation Results}
\label{Eva_EME}

\begin{table}[t]
	\centering
	\caption{3D object detection results on the VoD dataset quantified as mAP and sCD between the current point cloud and the ego-motion corrected point cloud from the previous time step. All results, irrespective of the ego-motion (EM) correction approach, are generated by applying PointPillars on radar data accumulated over 5 frames, ignoring the motion of dynamic objects. The best results are marked in \textbf{bold}. The arrow marks direction of better score.}
	\vspace{-0.125cm}
	\begin{tabular}{@{\extracolsep{1pt}}p{2.5cm}m{1.5cm}m{0.475cm}m{0.475cm}m{0.475cm}m{0.475cm}}
		\toprule   
		{} & {sCD [m] $\downarrow$} & \multicolumn{4}{c}{mAP $\uparrow$}  \\
		\cmidrule{3-6}
		\multirow{2}{2.5cm}{EM estimation approach}& {} & \multicolumn{2}{c}{3D} & \multicolumn{2}{c}{BEV} \\
		\cmidrule{3-4}
		\cmidrule{5-6}
		 &  & SR & LR & SR & LR \\ 
		\midrule
		No correction & 0.285 & 32.0 & 10.1 & 42.9 & 22.6 \\
		\midrule
		EM-P &	0.191  & 35.1 & 12.9 & 48.6 & 28.8 \\
		EM-G-Lidar &	\textbf{0.180}  & 38.4 & \textbf{14.2} & \textbf{50.5} & \textbf{29.6}  \\
		EM-mG-Lidar& 0.235  & 33.5 & 9.4  & 46.3 & 24.8  \\
		EM-G-Radar &	0.296  & 32.3 & 12.2 & 44.1 & 25.6  \\
		EM-$v_{rr}$	&0.235  & 32.9 & 11.9 & 45.6 & 27.7  \\
		EM-PCAc 	&0.216	  & \textbf{38.7} & 11.4 & 48.5 & 25.4 \\
		\bottomrule
	\end{tabular}
	\label{tab:comparison_em}
	\vspace{-0.25cm}
\end{table}

From Fig. \ref{tab:plot_em}, two categories of ego-motion patterns can be differentiated. Approaches resulting in a smooth motion estimation, EM-mG-Lidar and EM-$v_{rr}$, and algorithms with large deviations between frames. All deviating motion estimations show a similar pattern, most visible for the EM-G-Lidar estimated motion. This pattern repeats every 6 time steps. An explanation for this might be a time synchronization issue in data recording. This could also explain the apparent noise in the motion estimated from EM-P. Unfortunately, further investigations of this effect are impossible given the provided data. 
The amount of noise deviates between the different estimation strategies. Visual analysis shows that GICP-radar-based motion is most noise-prone. This can also be observed regarding the sCD between time steps in Tab. \ref{tab:comparison_em}. Employing EM-G-Radar for correction results in larger and therefore worse sCD compared to the uncorrected data. The EM-G-Radar reflects the time synchronization issue, and the inherent noisy measurements of the radar point cloud, which results in an insufficient correction in regard to the lidar point cloud, while still improving the object detection on radar data over no correction.
The robust point matching utilized in PCAc yields a better motion estimation from the radar point cloud. It has a significantly lower sCD and higher mAP for the SR 3D-mAP but cannot reach the EM-P and EM-G-Lidar estimation performance.
Comparing the sCD in Tab. \ref{tab:comparison_em} to the mAP shows that a low sCD corresponds to a better object detection performance.
All approaches improve on the uncorrected baseline. For EM-mG-Lidar and EM-$v_{rr}$, it is observable that the performance is low compared to the other approaches.
This strengthens the assumption that there is an inherent synchronization limitation.

\subsection{Influence of the Accumulation Horizon}
Different time horizons for point cloud accumulation are investigated to validate the aggregation with respect to the provided single frame data and to investigate the effects of longer time horizons for dynamic object handling. 
From Tab. \ref{tab:comparison_acc}, it is apparent that a large part of the improvement is already reached when accumulating just 3 frames. Accumulating more frames can result in better performance for some of the considered metrics but is diminished by worse performance in others. Accumulation over 5 frames balances good performance in SR and LR. This is especially apparent when the dynamic motion of objects is considered. A longer time horizon of 10 frames only improves the SR BEV performance, while in other cases, a shorter accumulation horizon results in better performance.

\begin{table}[t]
	\centering
	\caption{3D object detection results on the VoD dataset quantified as mAP. All models, irrespective of the accumulation horizon are generated by applying PointPillars on radar data corrected by the EM-P approach and if applicable, dynamic objects were corrected by the baseline method Dyn-GT. The best results, of only ego-motion and the best considering dynamic motion correction are marked in \textbf{bold}.}
	\begin{tabular}{@{\extracolsep{1pt}}p{4.0cm}p{0.475cm}p{0.475cm}p{0.475cm}p{0.475cm}}
		\toprule   
		{} & \multicolumn{2}{c}{3D} & \multicolumn{2}{c}{BEV} \\
		\cmidrule{2-3}
		\cmidrule{4-5}
		Accumulation strategy			 & SR & LR & SR & LR \\ 
		\midrule
		1 scan   & 30.0 & 9.5  & 42.7 & 22.1 \\
		\midrule
		3 scans   & \textbf{35.9} & 12.8 & 47.2 & 26.9 \\
		5 scans	 & 35.1 & \textbf{12.9} & 48.6 & \textbf{28.8} \\
		8 scans  & 35.7 & 12.0 & 49.1 & 26.4 \\
		10 scans  & 35.3 & 10.2 & \textbf{49.3} & 26.1 \\
		\midrule
		5 scans; Dyn-GT 	 & \textbf{36.3} & 13.5 & \textbf{47.6} & \textbf{27.1} \\
		10 scans; Dyn-GT    & 35.4 & \textbf{15.4} & 47.0 & 26.2 \\
		\bottomrule
	\end{tabular}
	\label{tab:comparison_acc}
	\vspace{-0.25cm}
\end{table}

\begin{figure}[t]
	\captionsetup[subfigure]{justification=centering}
	\centering
	\subfloat[Ground truth] {\includegraphics[height=4.3cm]{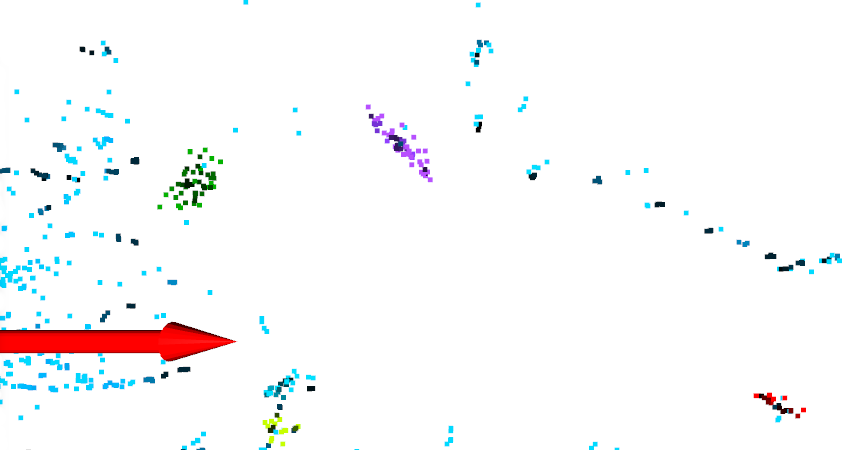}
		\label{Example_DME_GT}}
	\newline
	\vspace{0.2cm}
	\subfloat[Prediction]{\includegraphics[height=4.3cm]{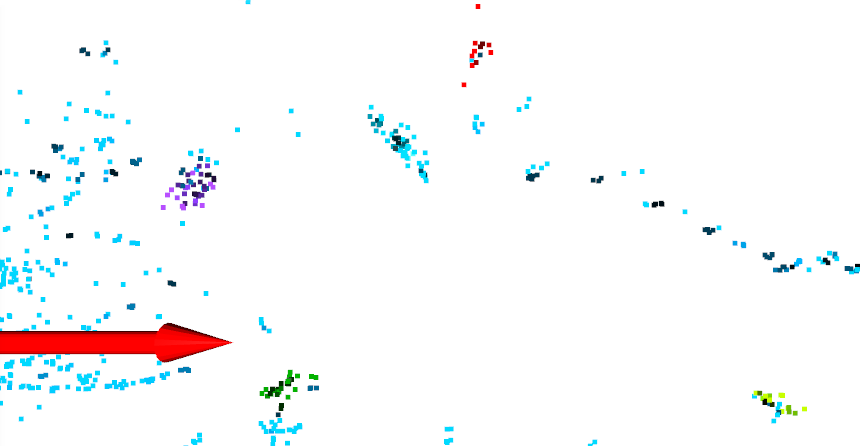}
		\label{Example_DME_Pred}}
	\caption{BEV perspective  of the instance association on an accumulated point cloud. The ego-vehicle is located at the origin of the red arrow (outside of image). Blue: static points; points in other colors than blue belong to object instances and the same color indicates the same instance ID.}
	\label{Example_DME}
	\vspace{-0.25cm}
\end{figure}

\subsection{Learning-Based Estimation of Dynamic Motion and Consideration for the Accumulation}

The evaluation in Sec. \ref{Eva_EME} has shown that the PCAc approach can estimate the ego-motion comparatively well, concluding that the PCAc approach works well on radar data too. Similar effects can be observed considering the instance association. Fig. \ref{Example_DME} shows that the model can distinguish between different instances, but not all objects are detected correctly. For example, the model could not estimate the purple-colored object in Fig. \ref{Example_DME_GT}. The dark green colored object in Fig. \ref{Example_DME_GT} has been estimated correctly, but not all of the points that belong to the object are correctly predicted as part of an object. Above the light green object in Fig. \ref{Example_DME_GT} is another object (observable in camera data) that is not labeled but predicted in Fig. \ref{Example_DME_Pred}. The red points in \ref{Example_DME_Pred} are predicted as dynamic while not belonging to any object. These effects could be a result of imperfect labels. Similar effects can be observed for the VoD lidar data.

Utilizing the object motion for the separate treatment of dynamic objects in accumulated point cloud for object detection shows a clear advantage for all approaches in Tab. \ref{tab:comparison_dyn} over just considering the ego-motion in Tab. \ref{tab:comparison_em}.
The naive approach of correcting the motion by just using $v_{rr}$ performs better than the label-based baseline. This could be due to errors in the motion estimation from the labels.
PCAc yields the best results. This is partly due to the inherently improved ego-motion estimation, while the estimation of dynamic motion improves on the results from Tab. \ref{tab:comparison_em} in all metrics but the SR 3D-mAP measurement.

\begin{table}[h]
	\centering
	\caption{
		3D object detection results on the VoD dataset quantified as mAP. All models, irrespective of the motion correction approach, are generated by applying PointPillars on radar data accumulated over 5 frames. The best results are marked in \textbf{bold}.}
	\vspace{-0.125cm}
	\begin{tabular}{@{\extracolsep{1pt}}p{4.0cm}p{0.475cm}p{0.475cm}p{0.475cm}p{0.475cm} }
		\toprule   
		\multirow{2}{3cm}{Dynamic motion estimation approach} & \multicolumn{2}{c}{3D} & \multicolumn{2}{c}{BEV} \\
		\cmidrule{2-3}
		\cmidrule{4-5}
		 & SR & LR & SR & LR  \\ 
		\midrule
		Dyn-GT & 36.3 & 13.5 & 47.6 & 27.1  \\
		Dyn-PCAc 	 & \textbf{38.7} & \textbf{16.0} & \textbf{51.4} & \textbf{31.1}  \\
		Dyn-$v_{rr}$	 & 37.6 & 13.8 & 50.1 & 29.0 \\
		\bottomrule
	\end{tabular}
	\label{tab:comparison_dyn}
	\vspace{-0.25cm}
\end{table}

\section{Conclusion}
\label{sec:conclusion} 
This work investigated the limitations of the VoD dataset for radar point cloud accumulation and approaches to mitigate missing ground truth ego-motion. It is shown that an improved ego-motion estimation improves the object detection performance. In addition to a better ego-motion estimation, a learning-based approach to mitigate the accumulation error for dynamic objects is applied to the dataset. Finally, it is shown that the motion of dynamic objects can be derived purely from the radar point cloud and that this improves object detection performance.

For future work, assuming a highly accurate inertial measurement-based ego-motion estimation will be available, the PCAc should be modified to utilize this motion instead of the estimated motion. This simplifies the architecture and might further improve the instance motion estimation. Furthermore, additional radar features like $v_{rr}$ and the radar cross-section could be integrated into PCAc for better instance motion estimation. This could also lead to improvements of the ultimate task of 3D object detection. Eventually, the PCAc model and the object detector could be trained as a single end-to-end model to boost the detection performance further.

\bibliographystyle{ieeetr}
\bibliography{ame_2023_references}

\begin{thebibliography}{10}

\bibitem{ViewOfDelftDataset_22}
A.~Palffy, E.~Pool, S.~Baratam, J.~F.~P. Kooij, and D.~M. Gavrila,
  ``Multi-class road user detection with 3+1d radar in the view-of-delft
  dataset,'' {\em IEEE Robotics and Automation Letters}, vol.~7, no.~2,
  pp.~4961--4968, 2022.

\bibitem{RadarSegmentation_18}
O.~Schumann, M.~Hahn, J.~Dickmann, and C.~Wöhler, ``Semantic segmentation on
  radar point clouds,'' in {\em 2018 21st International Conference on
  Information Fusion (FUSION)}, pp.~2179--2186, 2018.

\bibitem{FullVelocityRadarLong2021}
Y.~Long, D.~Morris, X.~Liu, M.~Castro, P.~Chakravarty, and P.~Narayanan,
  ``Full-velocity radar returns by radar-camera fusion,'' in {\em 2021 IEEE/CVF
  International Conference on Computer Vision (ICCV)}, pp.~16178--16187, 2021.

\bibitem{RigitSeneFlow_Dewan_2016}
A.~Dewan, T.~Caselitz, G.~D. Tipaldi, and W.~Burgard, ``Rigid scene flow for 3d
  lidar scans,'' in {\em 2016 IEEE/RSJ International Conference on Intelligent
  Robots and Systems (IROS)}, pp.~1765--1770, 2016.

\bibitem{Flownet3d_Liu_2019}
X.~Liu, C.~R. Qi, and L.~J. Guibas, ``Flownet3d: Learning scene flow in 3d
  point clouds,'' in {\em 2019 IEEE/CVF Conference on Computer Vision and
  Pattern Recognition (CVPR)}, pp.~529--537, 2019.

\bibitem{SSSS4DR_Liu_2022}
F.~Ding, Z.~Pan, Y.~Deng, J.~Deng, and C.~X. Lu, ``Self-supervised scene flow
  estimation with 4-d automotive radar,'' {\em IEEE Robotics and Automation
  Letters}, vol.~7, no.~3, pp.~8233--8240, 2022.

\bibitem{PointCloudAccumulation_22}
S.~Huang, Z.~Gojcic, J.~Huang, A.~Wieser, and K.~Schindler, ``Dynamic 3d scene
  analysis by point cloud accumulation,'' in {\em Computer Vision -- ECCV
  2022} (S.~Avidan, G.~Brostow, M.~Cisse, G.~M. Farinella, and T.~Hassner,
  eds.), (Cham), pp.~674--690, Springer Nature Switzerland, 2022.

\bibitem{TowarddeepradarZhou2022}
Y.~Zhou, L.~Liu, H.~Zhao, M.~López-Benítez, L.~Yu, and Y.~Yue, ``Towards deep
  radar perception for autonomous driving: Datasets, methods, and challenges,''
  {\em Sensors}, vol.~22, no.~11, 2022.

\bibitem{AstyxMeyer2019}
M.~Meyer and G.~Kuschk, ``Automotive radar dataset for deep learning based 3d
  object detection,'' in {\em 2019 16th European Radar Conference (EuRAD)},
  pp.~129--132, 2019.

\bibitem{VoD_Baratam_2022}
S.~Baratam, ``Radar-guided monocular depth estimation and point cloud fusion
  for 3d object detection,'' 2022.

\bibitem{GenerelizedICPSegal2009}
A.~Segal, D.~Hähnel, and S.~Thrun, ``Generalized-icp.,'' in {\em Robotics:
  Science and Systems} (J.~Trinkle, Y.~Matsuoka, and J.~A. Castellanos, eds.),
  The MIT Press, 2009.

\bibitem{AveragingQuaternionsMarkley2007}
F.~L. Markley, Y.~Cheng, J.~L. Crassidis, and Y.~Oshman, ``Averaging
  quaternions,'' {\em Journal of Guidance, Control, and Dynamics}, vol.~30,
  no.~4, pp.~1193--1197, 2007.

\bibitem{InstanteniousEgoMotionKellner2013}
D.~Kellner, M.~Barjenbruch, J.~Klappstein, J.~Dickmann, and K.~Dietmayer,
  ``Instantaneous ego-motion estimation using doppler radar,'' in {\em 16th
  International IEEE Conference on Intelligent Transportation Systems (ITSC
  2013)}, pp.~869--874, 2013.

\bibitem{RANSAC_Fischler_1981}
M.~A. Fischler and R.~C. Bolles, ``Random sample consensus: A paradigm for
  model fitting with applications to image analysis and automated
  cartography,'' {\em Commun. ACM}, vol.~24, p.~381–395, jun 1981.

\bibitem{Pointpillars}
A.~H. Lang, S.~Vora, H.~Caesar, L.~Zhou, J.~Yang, and O.~Beijbom,
  ``Pointpillars: Fast encoders for object detection from point clouds,'' in
  {\em 2019 IEEE/CVF Conference on Computer Vision and Pattern Recognition
  (CVPR)}, pp.~12689--12697, 2019.

\bibitem{openpcdet2020}
O.~D. Team, ``Openpcdet: An open-source toolbox for 3d object detection from
  point clouds,'' 2020.

\bibitem{Palmer_2023}
P.~Palmer, M.~Krueger, R.~Altendorfer, G.~Adam, and T.~Bertram, ``Reviewing 3d
  object detectors in the context of high-resolution 3+1d radar,'' in {\em
  Accepted for publication on 2023 IEEE/CVF Conference on Computer Vision and
  Pattern Recognition Workshop on 3D Vision and Robotics (CVPRW)}, pp.~1--10,
  2023.

\bibitem{OneCycle_Smith_2019}
L.~N. Smith and N.~Topin, ``{Super-convergence: very fast training of neural
  networks using large learning rates},'' in {\em Artificial Intelligence and
  Machine Learning for Multi-Domain Operations Applications} (T.~Pham, ed.),
  vol.~11006, p.~1100612, International Society for Optics and Photonics, SPIE,
  2019.

\bibitem{KITTI_Geiger_2013}
A.~Geiger, P.~Lenz, C.~Stiller, and R.~Urtasun, ``Vision meets robotics: The
  kitti dataset,'' {\em The International Journal of Robotics Research},
  vol.~32, no.~11, pp.~1231--1237, 2013.

\end{thebibliography}

\end{document}